\documentclass[a4paper]{article}

\newcommand{\documentdate}{\today}  

\usepackage{subfigure,graphicx}
\usepackage{a4wide,color}
\usepackage{natbib}

\usepackage{latexsym,amsmath,amstext} 
\usepackage[english]{babel}          
\usepackage[utf8]{inputenc}     
\usepackage{multirow}
\usepackage{multicol}
\usepackage{algorithm}
\usepackage{algorithmic}
\usepackage{bbm}
\usepackage{listings}
\usepackage{bbold}
\usepackage{ textcomp }

\definecolor{gray}{rgb}{0.4,0.4,0.4}
\definecolor{darkblue}{rgb}{0.0,0.0,0.6}
\definecolor{cyan}{rgb}{0.0,0.6,0.6}

\usepackage{amsmath}
\DeclareMathOperator*{\argmax}{arg\,max}

\def\R{\mbox{I\hspace{-.15em}R}}

\date{\documentdate}

\title{Comparison of Discrete Choice Models and Artificial Neural Networks in Presence of Missing Variables}

\author{
Johan Barthelemy$^\dagger$, Morgane Dumont$^\ddagger$ and Timoteo Carletti$^\ddagger$\\~\\
$^\dagger$SMART Infrastructure Facility, University of Wollongong, Australia\\
$^\ddagger$NaXys, University of Namur, Belgium\\~\\
johan\_barthelemy@uow.edu.au\\morgane.dumont@unamur.be\\timoteo.carletti@unamur.be}

\begin{document}

%
%

\maketitle

\begin{abstract}

Classification, the process of assigning a label (or class) to an observation given its features, is a common task in many applications.  Nonetheless in most real-life applications, the labels can not be fully explained by the observed features. Indeed there can be many factors hidden to the modellers. The unexplained variation is then treated as some random noise which is handled differently depending on the method retained by the practitioner. This work focuses on two simple and widely used supervised classification algorithms: \textit{discrete choice models} and \textit{artificial neural networks} in the context of binary classification.

Through various numerical experiments involving continuous or discrete explanatory features, we present a comparison of the retained methods' performance in presence of missing variables. The impact of the distribution of the two classes in the training data is also investigated. The outcomes of those experiments highlight the fact that artificial neural networks outperforms the discrete choice models, except when the distribution of the classes in the training data is highly unbalanced.

Finally, this work provides some guidelines for choosing the right classifier with respect to the training data.

\end{abstract}


\newpage
\section{Introduction and motivation}
\label{sec_introduction}

Classification, the process of assigning a label (or a class) to an observation given its features, is a common task in many applications and is often used in agent-based microsimulations. Indeed, the goal of those models being to simulate the behaviours and states of some generic interacting entities of interests called the agents, their core modules are designed to predict the actions made by the agents given their characteristics and their environment \citep{wooldridge2009introduction}.

This framework has been applied in countless applications such as 
\begin{itemize}
	\item transportation: selecting a path to reach a destination \citep{barthelemy2017adaptive}. Depending on the state of the traffic ahead of the agent and the time already spent travelling, the agent might reconsider its current path toward its destination;
	\item population dynamics: evolving a baseline synthetic population \citep{dumont2017importance}. Based on their characteristics, such as age, gender, position in the household and education level, the agents can decide to divorce, to find a spouse, to relocate,...;
	\item health: analysing the spread of Cholera in a population. Based on its information and believes about water quality and cost, the agents adapt their water consumption. \citep{Abdulkareem2018}.
\end{itemize}
Since the set of feasible actions are often finite and discrete, each action can be associated to a label and assigning the correct action becomes a classification problem\footnote{In this work, we will refer to label, decision, action, category and class indistinctly.}.

Many methods exist in the literature to perform classification tasks and can be divided into two categories: supervised and unsupervised. Supervised algorithms require labelled training data that also provides the class of each observation. This information is then used to train (or calibrate) the parameters of the models. On the other hand unsupervised algorithms do not have access to this information and try to divide the observations in homogeneous clusters.

It should be noted that in most real-life applications aiming to model and investigate complex agent-based systems, the decisions of the agents can not be fully explained by the observed variables or features. Indeed there can be many factors hidden to the modellers. The unexplained variation is then treated as some random noise which is handled differently depending on the method retained by the practitioner.

This work focuses on two simple and widely used supervised algorithms applied to predict categories or choices made by agents: \textit{discrete choice models} and \textit{artificial neural network}. More specifically, we will discuss the performance of the multinomial logit discrete choice model against a shallow feed forward artificial neural network. Those two methods have already been compared in previous works related to mode choice \citep{hensher2000comparison}, transport demand forecasting \citep{de1998forecasting}, driver compliance with traffic information \citep{dia2010evaluation}, hydrogeology \citep{barthelemy2016interaction} and population dynamics \citep{dumont2017robustness}.

Nonetheless, to the best of the authors' knowledge, a comparison of their performance in presence of missing variables has not been done yet, thus motivating this study. In addition, in many relevant applications, the features are unevenly shared among the classes. We are thus also interested in studying the impact of the distribution of the various categories in the training data as this can have a significant impact on the training of the algorithms and the resulting outcomes.

The remainder of this exploratory paper is organised as follows. The multinomial logit and the feed forward artificial neural network as well as the data used for training the algorithms are firstly described in Section 1. The numerical experiments comparing accuracy of the methods with respect to the number of missing variables and the distribution of the categories in the training data are then presented in Section~2. Concluding remarks and perspectives are presented in the last section.

\section{A brief overview of the methods and the data}
\label{sec_models_data}

This Section briefly introduces the two well known methods that will be compared in this work, namely the multinomial logit discrete choice model (DC) and the feed forward artificial neural network (ANN). Let us denote by $\mathcal{X}$ and $\mathcal{C}$ the set of observations and categories, respectively. Those models can both be used to determine the probability $p_i$ that a given observation $x \in \mathcal{X}$ belongs to a category $c_i \in \mathcal{C}$ for $i=1...m$ with $m$ being the number of categories that it is assumed to be known. More formally, they can be written as a mapping $f$:
\begin{equation}
x \in \mathcal{X},~\mathcal{C} = \{c_1, \dots, c_m \} \quad \leadsto \quad p =  f(x ~ | ~ \mathcal{C}, \Theta) \in \R^m 
\end{equation}
such that $\sum_i p_i = 1$, $0 \leq p_i \leq 1$ and where $\Theta$ is the set of parameters of the model that need to be estimated. 

The probabilities $p_i$ can be interpreted as the likelihood or confidence of a given observation belonging to each category $c_i$. The predicted probabilities can be converted into a class value $c_k$ by selecting the class label that has the highest probability, i.e. $k = \arg\max_i p_i$. For example, let us assume that we have two classes $c_1$ and $c_2$ with predicted probabilities $p_1=0.62$ and $p_2=0.38$ for a given observation $x$. Then, $x$ will be labelled $c_1$ since $p_1>p_2$. 

Before moving to the description of the methods and the estimation of their parameters, we first detail the datasets used for training and validating the methods.

\subsection{Training and validation datasets}
\label{datasets}

The dataset used in this work have been artificially generated thanks to the method \verb?make_classification? of the \verb?scikit-learn? module for the Python 3 programming language \citep{scikit-learn}. This method adapts the algorithm that was originally designed to generate the MADELON dataset \citep{guyon2003design}.

The resulting dataset contains 5,000 observations $x_i$, each of them described by 100 features $v_j$ ($j = 1,\dots,100$) randomly drawn following a standard Normal distribution, i.e. $v_j \sim \mathcal{N}(0,1)$, and one binary dependent variable $y \in \mathcal{C}=\{0,1\}$ being the class of the observation. Each feature $v_j$ has the same scale and is informative, i.e. there is no redundant or duplicated variables in the dataset. Some noise is also introduced by randomly flipping the value of $y$ for one percent of the observations. We will denote by $X = [X_1,\dots,X_{100}]$ the matrix of dimension $5000 \times 100$ containing the data. Each column and row corresponds to a feature and an observation, respectively.

Initially, the distribution of the two classes defined by $\mathcal{C}$ is balanced, i.e. the proportion $p_y$ of observations for which $y=0$ is $50\%$. Later on, the proportions $p_y \in \{0.6, 0.7, 0.8, 0.9\}$ will also be tested.

In the numerical experiments, an alternative version of the dataset in which the features are discrete will also be considered. Those new discrete features $X_i'$ of the matrix $X'$ are obtained by applying the following transformation to each column of $X$ (where $\lfloor \bullet \rfloor$ is the flooring operator):
\begin{equation}
X_i' = \lfloor X_i \times 5 + 0.5 \rfloor.
\end{equation}

Finally, the dataset is randomly split into a training and a validation datasets: $75\%$ of the observations will be used for the training and the remaining $25\%$ will be used for the validation.

\subsection{Logit discrete choice model}

Discrete choice models aim to explain and predict choices amongst a finite and exhaustive set of mutually exclusive alternatives $\mathcal{C}$ \citep{ben1985discrete}.

The key assumption of this methodology is that the agent will always opt for the alternative that maximises its utility (or benefit). Let us denote by $U_{xy}$ the utility that the agent $x$ associates with the alternative $y \in \mathcal{C}$. Then, the agent will choose the alternative $y^\star$ if
\begin{equation}
U_{xy^\star } > U_{xy} \qquad \forall y \in \mathcal{C} \text{ and } y \neq y^\star.
\end{equation}

Typically, the choice depends on many factors and some of those may remain unknown for the observer. The utility $U_{xy}$ perceived by the agent $x$ for the alternative $y$ can be divided into two parts:
\begin{equation}
U_{xy} = V_{xy} + \epsilon_{xy}
\end{equation}
where $V_{xy}$ and $\epsilon_{xy}$ respectively denote the observed and hidden parts of the utility. The probability that an agent $x$ retains the alternative $y^\star$ is then given by
\begin{eqnarray}
  P_{xy^\star } & = & P(U_{xy^\star} > U_{xy}, \quad \forall y \neq y^\star)\\
         & = & P(\epsilon_{xy} - \epsilon_{xy^\star } < V_{xy} - V_{xy^\star}, \quad \forall y \neq y^\star).
\end{eqnarray}
Those two equations highlight two important characteristics of discrete choice models: only the difference in utility matters\footnote{Not the absolute value.} and the overall scale of the utilities is irrelevant\footnote{Multiplying each utility by the same factor $\alpha$ does not change the ordering of the utilities}. The former implies that one alternative can have a utility set to 0 and is referred to as the base alternative.

In our context, we assume the observed utility to have the form of a linear model
\begin{equation}
V_{xy} = \beta_y^T x
\label{eq_obs_utility}
\end{equation}
where $x$ and $\beta_y$ are two vectors, the former containing the values of the observed variables for the agent $x$ and the latter containing the model coefficients (or parameters).


Different specifications of the random components $\epsilon_{xy}$ lead to different models. By assuming that those terms are independent and follow an identical standard Gumbell distribution whose cumulative distribution function is defined by:
\begin{equation}
F(\epsilon_{xy}) = e^{-e^{-\epsilon_{xy}}},
\end{equation}
where $e$ is the Euler's constant, it can be shown that the probability associated with each alternative is given by \citep{ben1985discrete}:
\begin{equation}
P_{xy} = \frac{e^{V_{xy}}}{\sum_k e^{V_{xk}}}
\end{equation}

The maximum likelihood approach can be used to obtain the vectors $\hat{\beta}_y$ estimating the vectors of coefficients $\beta_y$ in Equation~\ref{eq_obs_utility}:
\begin{equation}
\hat{\beta} = \argmax_{\beta} \mathcal{L}(\beta) = \sum_i \sum_y  \mathbb{1}_{iy} \left( \beta_y^T x_{i} - \ln \sum_{z} e^{\beta_{z}^T x_{i}} \right)
\end{equation}
where $\mathbb{1}_{iy} = 1$ if $x_i$ chooses the alternative $y$ in the training data and $0$ otherwise. The function \verb&MNLogit& of the Python 3 module \verb&statsmodel& \citep{seabold2010statsmodels} has been used to perform the estimation.

\subsection{Feed forward artificial neural network}

Artificial neural networks are a class of supervised machine learning algorithms inspired by the biological neural networks: the neurons receive a stimuli (or input) from previous neurons, process it and forward the outcome to the following neurons if it is strong enough. It has been shown that neural network can approximate any function \citep{cybenko1989approximation,hornik1990universal}, hence this approach has been used in countless application, including classification. An extensive introduction to this methodology can be found in \citep{kriesel2007introduction} and \citep{ripley2007pattern}.

\begin{figure}[h!]
\begin{center}
\includegraphics[width=0.55\textwidth]{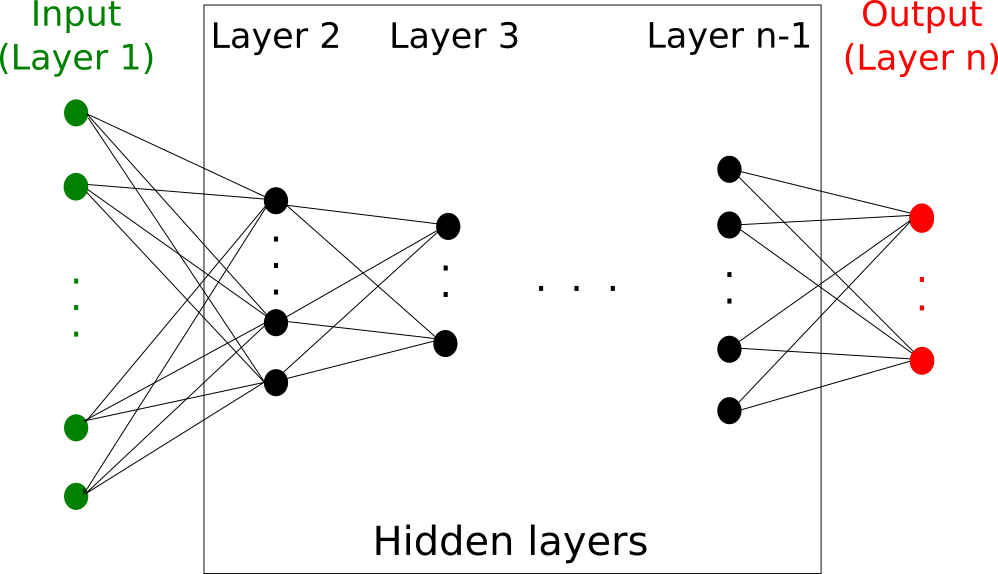}
\caption{Illustration of the architecture of a neural network composed by $n$ fully interconnected layers. The neurons in the first layer (in green) receive the inputs and are connected with every neuron of the second layer. The information progress from each layer $k$ to the next one $k+1$. The last layer (in red) contains the outputs of the neural network. Between the input and output layers, there are $n-2$ hidden layers.}
\label{fig:Illus_NN}
\end{center}
\end{figure}

Typically an artificial neural network is organised in interconnected layers of neurons. More specifically the feed forward architecture is characterised by one input layer, one or more hidden layers and one output layer.  This architecture can be described by an array $v = [v_1,...,v_n]$ where each component $v_i$ corresponds to the size of the layer $i$, i.e, the number of neurons in that layer, $v_1$ corresponds to the input layer and $v_n$ to the output layer. Furthermore each neuron in a layer $k$ is connected to each neuron in layer $k+1$ as illustrated in Figure~\ref{fig:Illus_NN}. 

Each neuron $o_j^{k+1}$ in a given layer $k+1$ first transforms the values received by the neurons  $o_i^k$ in the previous layer with a weighted sum resulting in a temporary value:
\begin{equation}
t_j^{k+1} = \sum_{i = 1}^{v_k} w_{i,j}^k \times o_i^k + \theta_{j}^{k+1}
\end{equation}
where $w_{i,j}^k$ are the weights of the connections of the neurons $i$ (in layer $k$) with the $j$ one (in layer $k+1$) and the term $\theta_{j}^{k+1}$ is the bias added to the neuron $o_j^{k+1}$.
This summation is then followed by a non-linear activation function $g$ to obtain the final value at $o_j^{k+1}$. In our context, $g$ is the \emph{ReLU} and $o_j^{k+1}$ is then given by:
$$
o_j^{k+1} = g(t_j^{k+1}) = \max\{0, t_j^{k+1} \}.
$$

Each neuron $o_i^n$ of the output layer computes the probability that the observation given to the input layer belongs to the category $i$. In order to compute the probability that a given input belongs to the category $i$, the softmax function is applied to the neurons in output layer which is defined by:
\begin{equation}
p_i = \frac{e^{o_i^n}}{\sum_k e^{o_k^n}}.
\end{equation}

The values of the weights and the biais are determined by minimising a loss function. The specification of the function depends on the task to be performed by the artificial neural network. In the case of classification, the loss function is given by the cross-entropy defined by:
\begin{equation}
L(\hat{p},c) = - c \ln \hat{p} - (1 - c) \ln(1 - \hat{p}) 
\end{equation}
where $\hat{p}$ is the vector of predicted probabilities and $c$ is an indicator vector where $c_i = 1$ if the observation belongs to the category $i$ and $0$ otherwise.

It can be noted that compared to the discrete choices models, this method belongs to the framework of deterministic classifiers, because the noise is not explicitly taken into account\footnote{Even if one can add an extra input variable whose values are randomly drawn from some distribution to mimic the presence of unknown.}.

In the experiments illustrated in the next Section, we will consider a simple feed forward architecture $[j,15,5,2]$ with $1 \leq j \leq 100$. The Python 3 module \verb&scikit-learn& has been used to optimise the parameters of the neural network.

\section{Numerical experiments}
\label{sec_results}

This Section investigates the prediction performance of the ANN and DC methods when the proportion of used features evolves from 0.01 to 1.00, thus simulating scenarios with different proportion of missing variables. 

In order to assess the quality of the models' predictions, we need to use some metrics. For a binary classification task, those measures will be extracted from the confusion matrix illustrated in Table~\ref{tab:confusion}. As there are only two classes, they can be labelled \emph{True} and \emph{False}. The sum of the elements of this matrix corresponds to the total number of predictions.
\begin{table}[h!]
\begin{center}
\begin{tabular}{|l|c|c|}
\hline
ground truth \textbackslash ~ predicted & False & True\\
\hline
\hline
False & True Negative (TN) & False Positive (FP)\\
\hline
True & False Negative (FN) & True Positive (TP)\\
\hline
\end{tabular}
\caption{Confusion matrix}
\label{tab:confusion} 
\end{center}
\end{table}

We will consider the four indicators detailed below:
\begin{enumerate}
	\item the \emph{accuracy} being the proportion of observations predicted in the correct class:
$$Accuracy =  \frac{TN+TP}{TN+TP+FN+FP}$$	
	\item the \emph{precision} calculated as the ratio of the number of true positive over the number of positive predictions:
	$$Precision = \frac{TP}{TP+FP} $$
	\item the \emph{recall} defined as the ratio of the number true positive over the total of observations that are actually positive.
	$$Recall = \frac{TP}{TP+FN} $$
	\item and the harmonic mean of the \emph{recall} and the \emph{precision} denoted \emph{F1}:
	$$F1 = \frac{2*Recall*Precision}{Recall + Precision}.$$
\end{enumerate}

In addition, two types of probabilities are also taken into account: the probability of the actual class (in the training data) and the probability of the predicted class. The latter can be seen as the level of certainty of the algorithm. These probabilities being defined for each objects of a simulation. For example, let us consider an observation $x$ belonging to the class \textit{True}. According to the classifier, the probability of $x$ being \textit{True} is $0.46$ while the predicted label is \textit{False} with probability $0.54$.

Now that we have defined our performance quantifiers, we can turn to present the numerical experiments, divided into two sets. The first one assumes that the observations are uniformly distributed amongst the two classes, while the second set will test several proportions for the distribution of the observations within the two classes. In each set, the explanatory variables will be either continuous or discrete.

To investigate the impact of different proportion of missing variables, the models will be trained 20 times.

\subsection{Balanced classes}

In these first experiments, the two clusters are balanced, meaning that half of the observations belong to each class. We first examine the case where the features are continuous before analysing the discrete case.

\subsubsection*{Continuous variables}

As mentioned previously, for each number of used variable $v$, we train the models 20 times using $v$ variables randomly chosen without replacement.

\begin{figure}[h!]
\begin{center}
\includegraphics[width=0.65\textwidth]{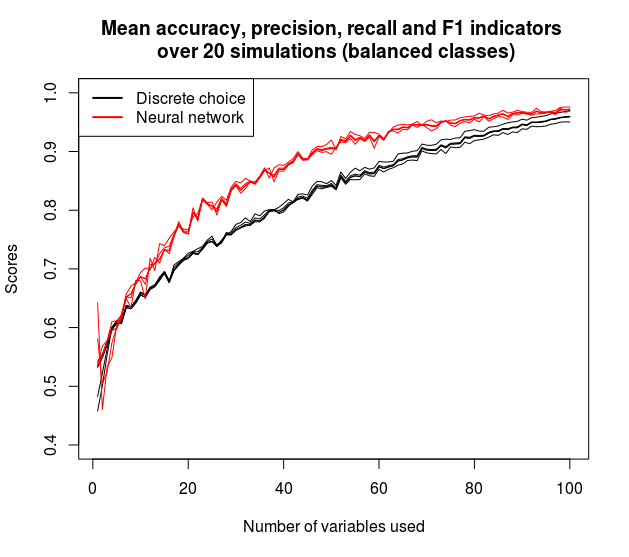}
\caption{The mean scores (Accuracy, Precision, Recall, F1) per number of included variables over 20 simulations when features are continuous. For both methods, the scores increase when adding variables, with the neural network early performing better than discrete choice modelling.}
\label{fig:ScoresRealBalanced}
~\\
\includegraphics[width=0.65\textwidth]{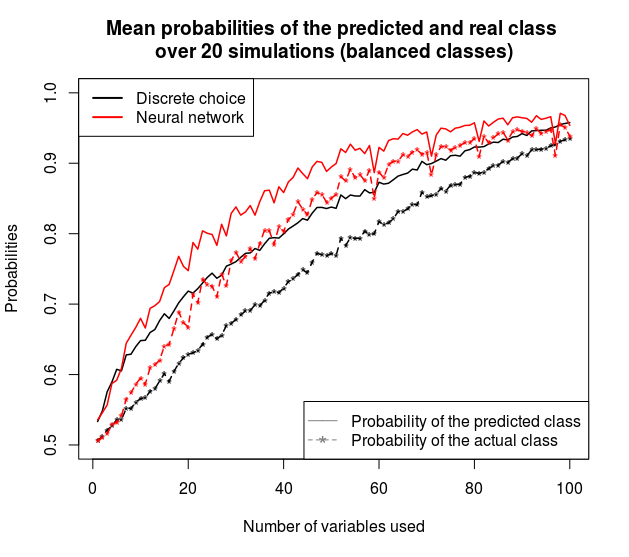}
\caption{The mean probabilities of the predicted and actual classes per number of included variables over 20 simulations when features are continuous. For both methods, the probabilities increase when adding variables, with the neural network performing better than discrete choice modelling.}
\label{fig:ProbaRealBalanced}
\end{center}
\end{figure}

Figure~\ref{fig:ScoresRealBalanced} shows the evolution of the mean\footnote{computed over the 20 runs on the validation dataset} of the four scores for each method against the number of retained variables $v$. For a sake of readability and because the trends of the 4 measures are very close, we didn't present an extended legend. We can observe that with less than 10 variables, both models have similar poor scores. As  $v$  is increasing, we note an improvement in the predictions, which is steeper for the neural network. Once $v \geq 50$, then the gap between the methods tightens. Eventually, when 100\% of the variables are used, both methods reach mean scores above 0.9. In summary, for this particular configuration of balanced classes, the artificial neural network performs better than the discrete choice models, being always above the discrete choice method.

The evolution of the mean\footnote{computed over the 20 runs and every observation in the validation dataset} probabilities computed by the models for the true and predicted classes is shown in Figure~\ref{fig:ProbaRealBalanced}. It is clear that probabilities increase with the number of variables. The neural network approach is obviously outperforming the discrete choice approach, even though both have the same performance in two cases: when there is no missing variables; and when the number of used variable is less than 5.

\clearpage
In summary, when the classes are balanced and in presence of missing variables, the neural network approach seems to be the best option. As expected, the performance of the neural network and the discrete choice model increase when the number of missing variable is decreasing.

\subsubsection*{Discrete variables}

\begin{figure}[h!]
\begin{center}
\includegraphics[width=0.65\textwidth]{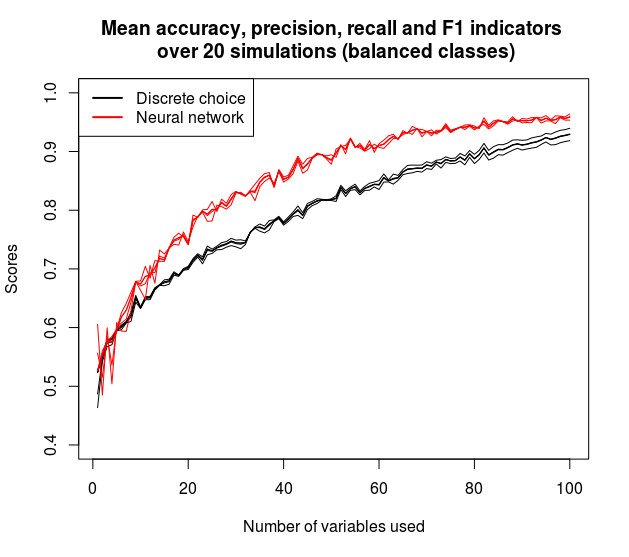}
\caption{The mean scores (Accuracy, Precision, Recall, F1) per number of included variables over 20 simulations when features are discrete. For both methods, the scores increase when adding variables, with the neural network early performing better than discrete choice modelling.}
\label{fig:ScoresIntegerBalanced}
\end{center}
\end{figure}

\begin{figure}[h!]
\begin{center}
\includegraphics[width=0.65\textwidth]{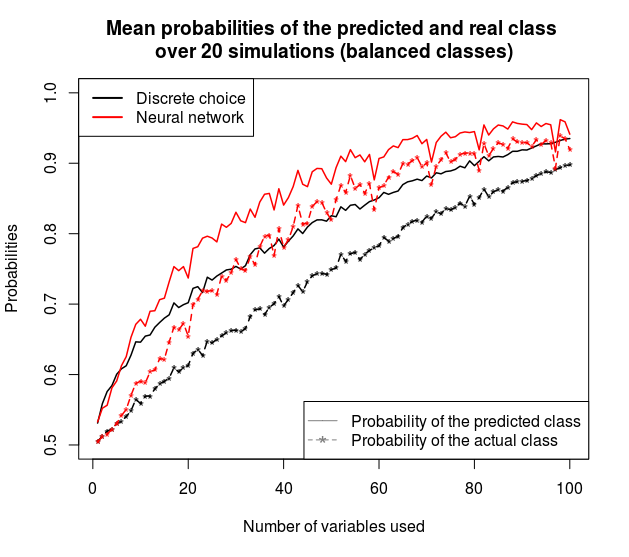}
\caption{The mean probabilities of the predicted and actual classes per number of included variables over 20 simulations when features are discrete. For both methods, the probabilities increase when adding variables, with the neural network performing better than discrete choice modelling.}
\label{fig:ProbaIntegerBalanced}

\end{center}
\end{figure}

We now investigate whether using discrete explanatory variables influence the choice of the best prediction method. The results are displayed in Figures~\ref{fig:ScoresIntegerBalanced} and~\ref{fig:ProbaIntegerBalanced}. One can easily observe that the behaviour of the indicators and the probabilities are similar to the ones when the variables are continuous. Hence the conclusion remains the same and the neural network approach should be again favoured.

\clearpage
\subsection{Unbalanced classes}

We have seen previously that the number of missing variables influence the prediction performance of the methods when the observation are uniformly distributed within the two classes. In this Section we now experiment different allocation of the observations amongst the two classes. Let us denote by $p_T \in ]0,1[$ and $p_F = 1 - p_T$ the proportion of observations in the class \emph{True} and \emph{False}, respectively. We will test the following values for $p_T$:
$$
0.5,~0.6,~0.7,~0.8~\text{and } 0.9.
$$

\subsubsection*{Continuous variables}

\begin{figure}
\begin{center}
\includegraphics[trim=2cm 10cm 2cm 11.5cm,clip=true,width=\textwidth]{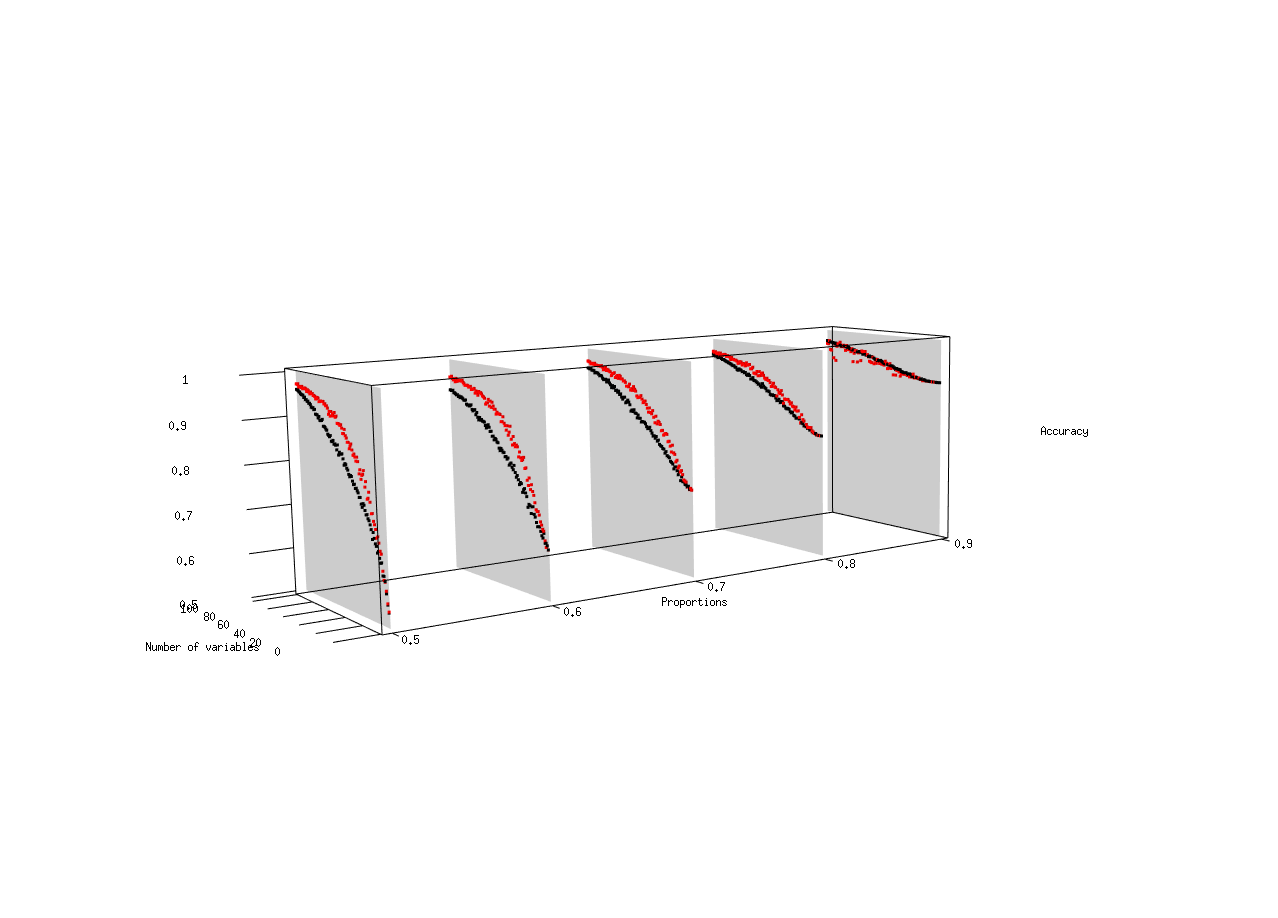} \\ (a) Accuracy
\\\includegraphics[trim=2cm 10cm 2cm 11.5cm,clip=true,width=\textwidth]{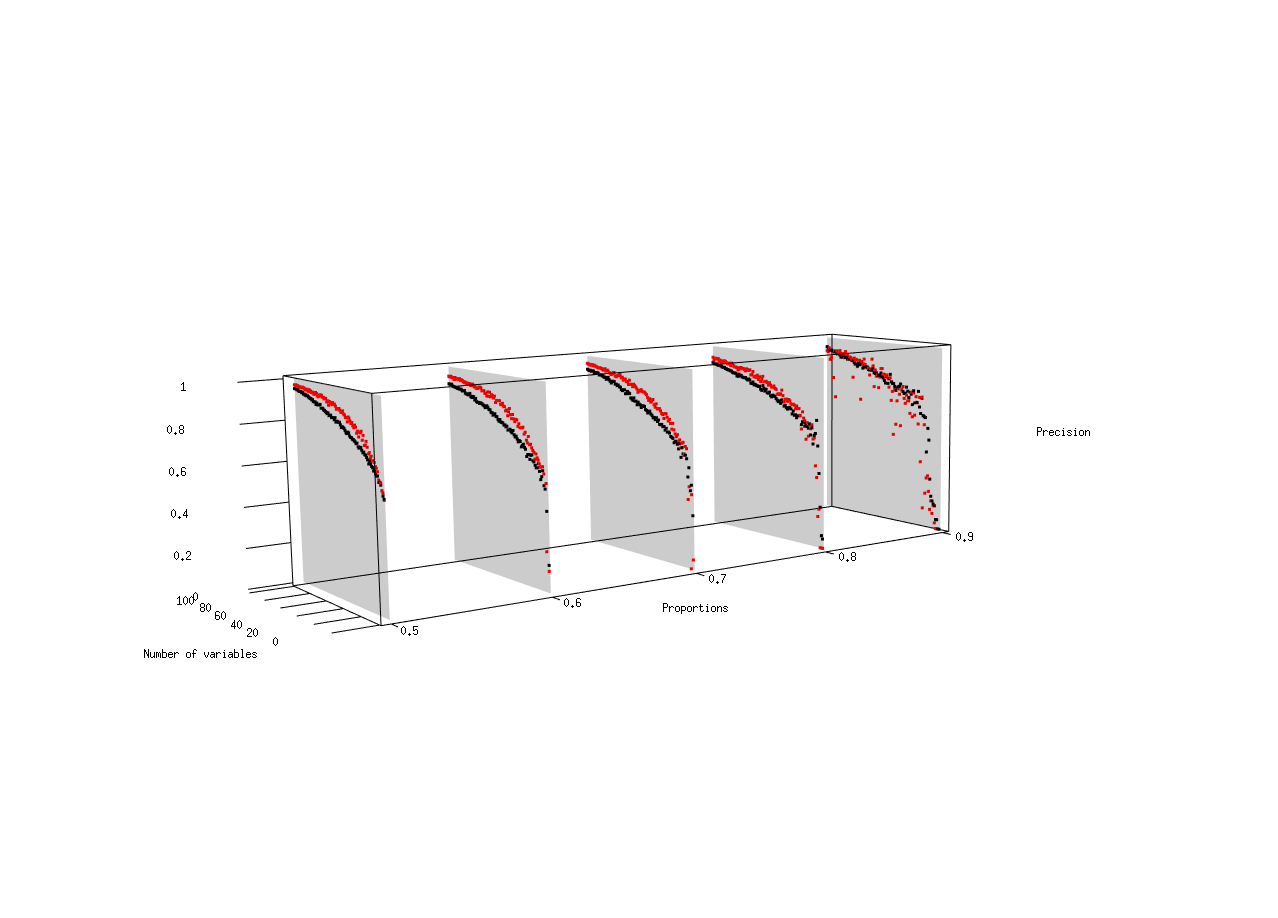}\\ (b) Precision\\
\includegraphics[trim=2cm 10cm 2cm 11.5cm,clip=true,width=\textwidth]{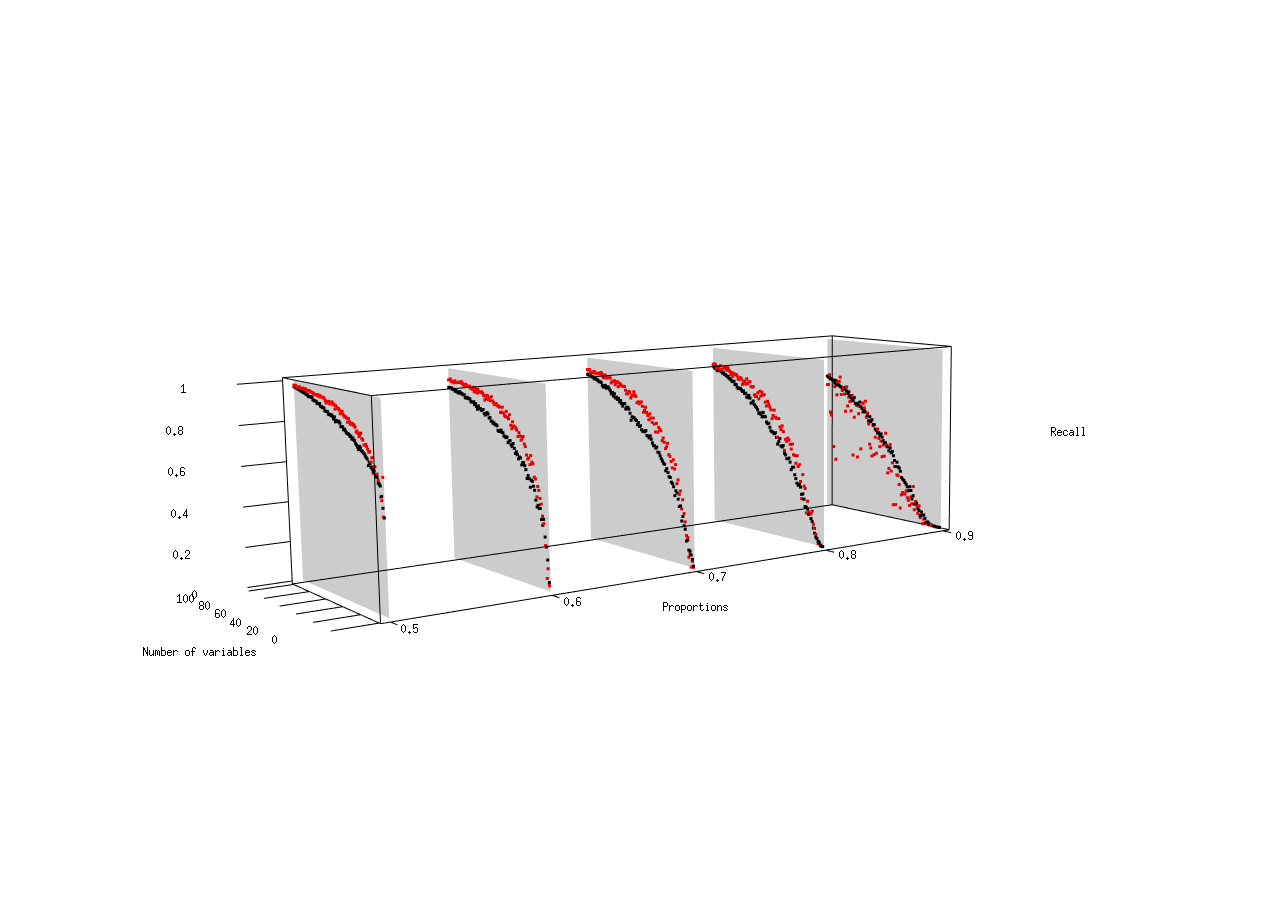}\\ (c) Recall
\\\includegraphics[trim=2cm 10cm 2cm 11.5cm,clip=true,width=\textwidth]{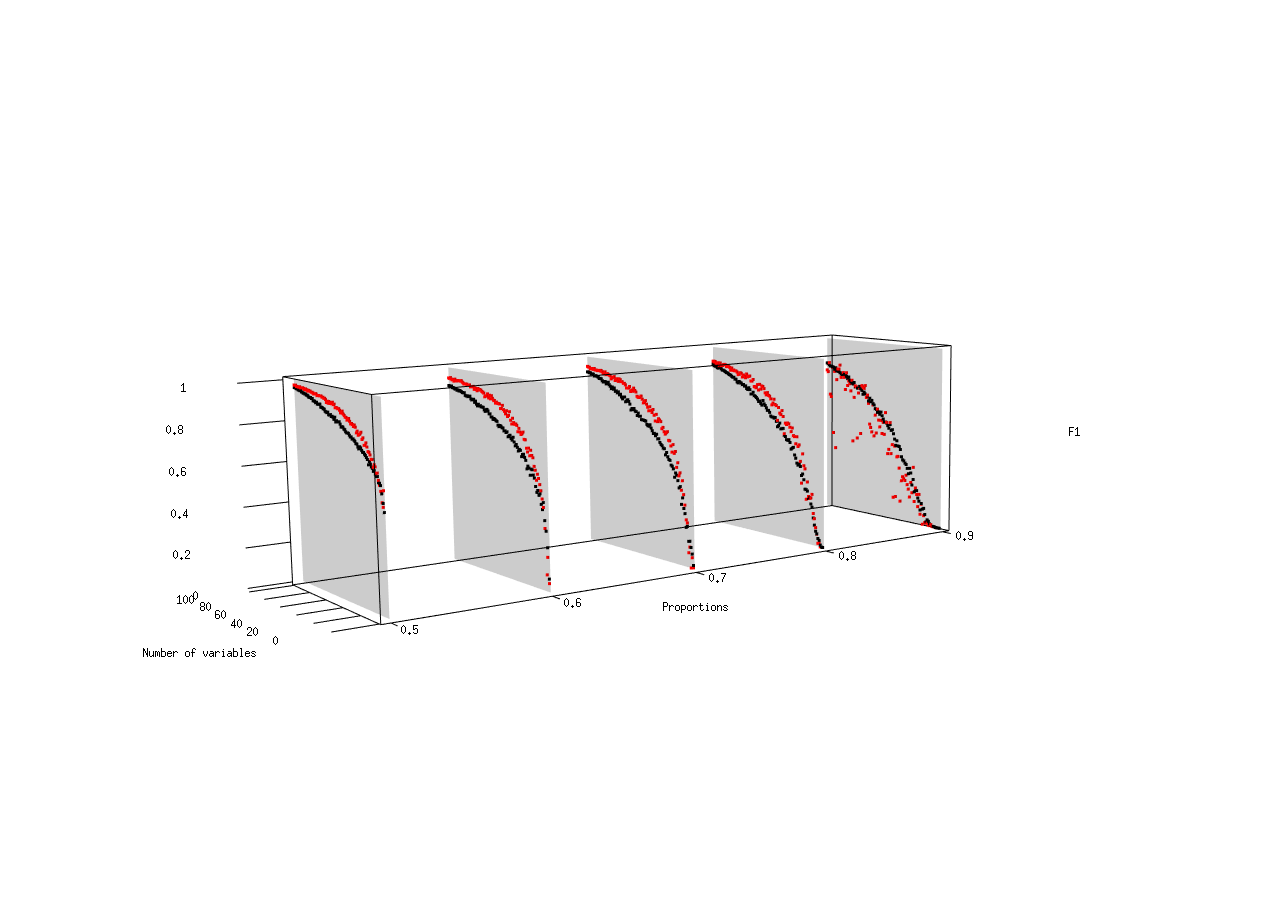} \\ (d) F1\\
\caption{The mean accuracy (a), precision (b), recall (c) and F1 (d) scores per number of included variables and $p_T$ over 20 simulations when features are continuous. It can be seen that only when $p_T = 0.9$, the discrete choice model (black dots) performs better than the artificial neural network (red dots). For the other proportions, the neural network generates better predictions. It can also be seen that more variables results usually in better predictions exception for the neural network when $p_T = 0.9$.}
\end{center}
\label{fig:ScoresRealImbalanced}
\end{figure}

We first look at the performance of the methods when the explanatory features are represented by continuous variables. The results in terms of the four scores are illustrated in Figure~\ref{fig:ScoresRealImbalanced}. An interactive version of these 3D plots is available at \emph{https://plot.ly/$\sim$modumont/11}. The first observation is that the accuracy for models with only one variable is close to the proportion $p_T$, meaning that this score is directly influenced by the amplitude of unbalanced classes. This is intuitive, since when classes are unbalanced, if a model associates all observations to the dominant class, then the accuracy is $p_T$. For all other scores, models considering only one variable implies low quality. Secondly, the proportion $p_T$ seems to have a different impact on both methods. Indeed, artificial neural network (red dots) results in fuzzy  scores when the sample is really unbalanced ($p_T=0.9$), with simulations performing sometimes better with less variables. However, discrete choice models (black dots) are systematically improved when the number of used variables increases, for each proportion $p_T$. These experiments show that artificial neural network are more efficient when having balanced or slightly unbalanced classes, whereas it becomes unstable with highly unbalanced classes. Discrete choice models stays stable and take advantage of additional variables in all cases of unbalanced classes.


\begin{figure}
\begin{center}
\includegraphics[trim=2cm 9cm 2cm 10cm,clip=true,width=\textwidth]{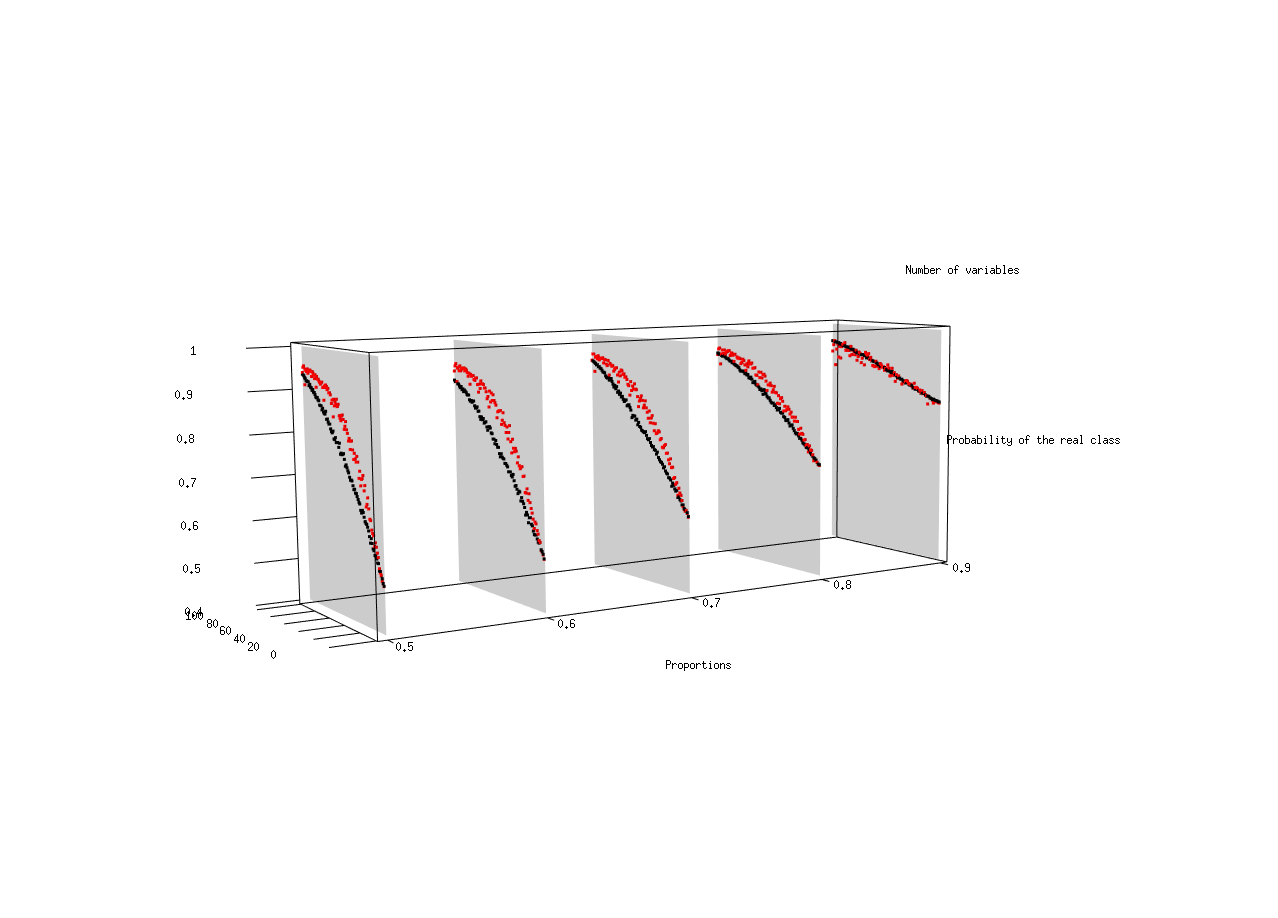}\\ (a) actual  \vspace{2cm}\\ \includegraphics[trim=2cm 9cm 2cm 10cm,clip=true,width=\textwidth]{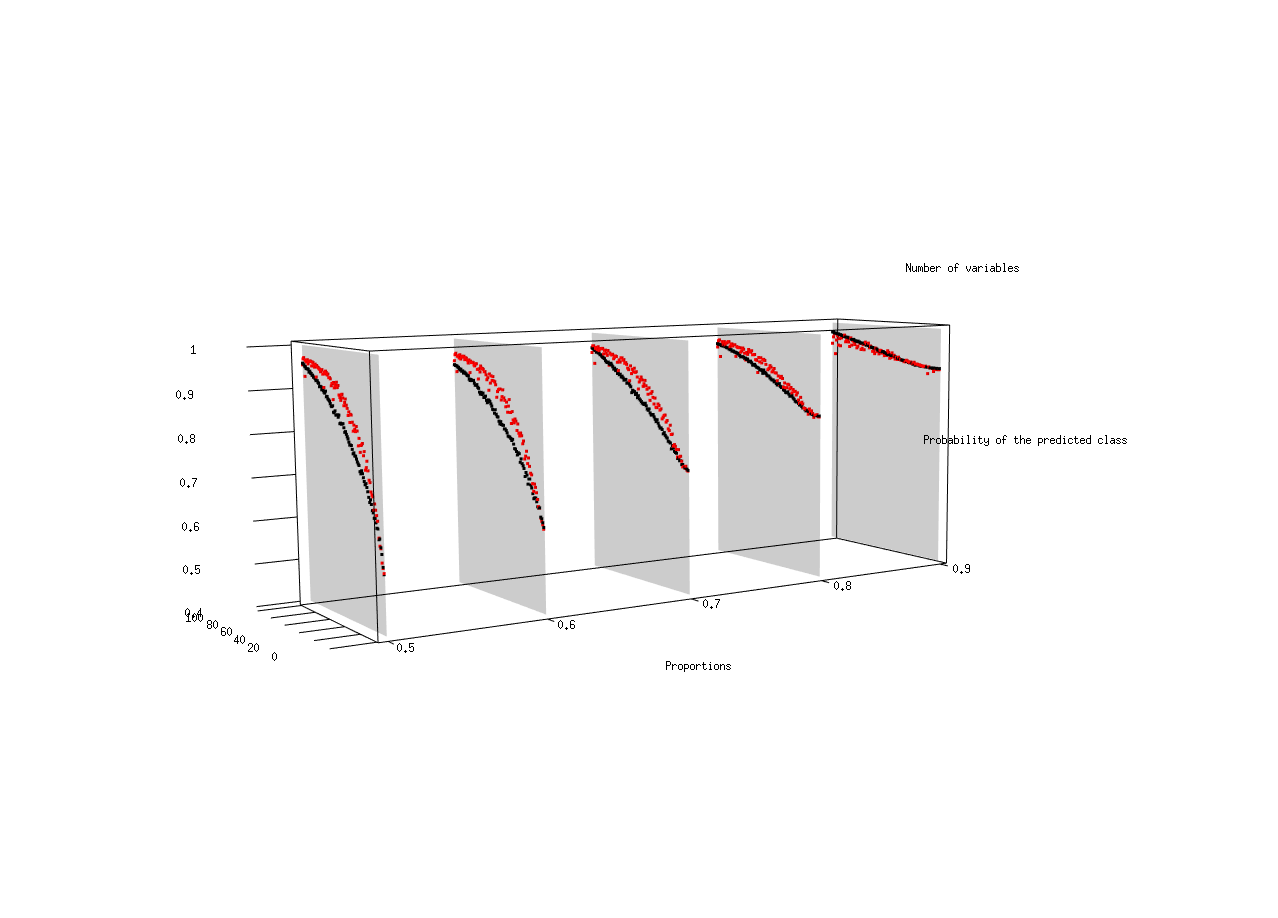} \\(b) predicted
\caption{Probabilities of the actual (a) and predicted (b) classes generated by the artificial neural network (red) and the discrete choice model (black) for continuous variables. The artificial neural network has more confidence into its predictions, since the probabilities are higher (panel b) except for $p_T = 0.9$. In addition, the same can be seen for the predicted probabilities of the actual class. When $p_T=0.9$, the situation is a bit fuzzy and there is no clear winner, even though the DC seems to be more consistent. }
\end{center}
\label{fig:ProbaRealImbalanced}
\end{figure}

It can also be seen on Figure~\ref{fig:ProbaRealImbalanced}, displaying the probabilities of predicted and actual classes, that discrete choice models also result in smoother increases with respect to the number of used variables. Furthermore, the conclusions are similar to the ones given above regarding the scores. An interactive version of these 3D plots is available at \emph{https://plot.ly/$\sim$modumont/15/}.

In summary, if classes are highly unbalanced ($p_T$=0.9), the safer choice is to use a discrete choice model, whereas in all other configurations, artificial neural networks would be a better option. In real life, such situations occur for example in the context of credit card fraud detection \citep{fiore2017using} or modelling the marriages within a synthetic population \citep{dumont2017robustness}.

\subsubsection*{Discrete variables}

As for the balanced case, the results of the experiments relying on discrete features are similar to the ones obtained with continuous features. Hence they will not be presented here in detail, but the interested reader can find the graphs in Appendix.
\clearpage
\section{Conclusions}
\label{sec_conclusions}

This work has investigated the impact of two factors on the prediction performance of the Logit discrete choice model and the artificial neural network for a binary classification task: the number of missing variables simulating imperfect information and the distribution of the individuals amongst the two classes. 

The numerical experiments indicate that the neural network should be favoured most of the time, expect when the classes are highly unbalanced. Indeed, when 90\% of the observations belongs to the same class, the discrete choice models seems to be a better option. 

In addition, we have also compared the use of discrete and continuous explanatory variables and the conclusions remain the same in both cases.

As these kind of situations arise quite often, this work can be of interests for researchers wondering which of these methods to chose for all problems equivalent to a binary classification. Indeed, by checking how imbalanced the data is, the practitioner can then select the right approach. 

This work could be extended by considering more classes, additional supervised classification methods (Logit regression, Support Vector Machine, decision tree ...), adapting the structure of the ANN, using another discrete choice model (such as Probit)  and testing different datasets.

{\footnotesize
\section*{\footnotesize Acknowledgements}
The authors wishes to thank Gauthier Sanzot for his helpful corrections and suggestions. Computational resources have been provided by the Consortium des Équipements de Calcul Intensif (CÉCI), funded by the Fonds de la Recherche Scientifique de Belgique (F.R.S.-FNRS) under Grant No. 2.5020.11.  Finally we gratefully acknowledge the support of NVIDIA Corporation with the donation of the Titan Xp GPU used for this research.
}

\appendix
\section*{Appendix}

The results for different proportions $p_T$ of observations belonging to the class \emph{True} and characterized by discrete features are illustrated in the figures below. Those results are similar to the ones obtained when using continuous features. Interactive versions of these 3D plots are available at \emph{https://plot.ly/$\sim$modumont/13/} for the scores and \emph{https://plot.ly/$\sim$modumont/17/} for the probabilities.

\begin{figure}[h!]
\begin{center}
\includegraphics[trim=2cm 10cm 2cm 11.5cm,clip=true,width=\textwidth]{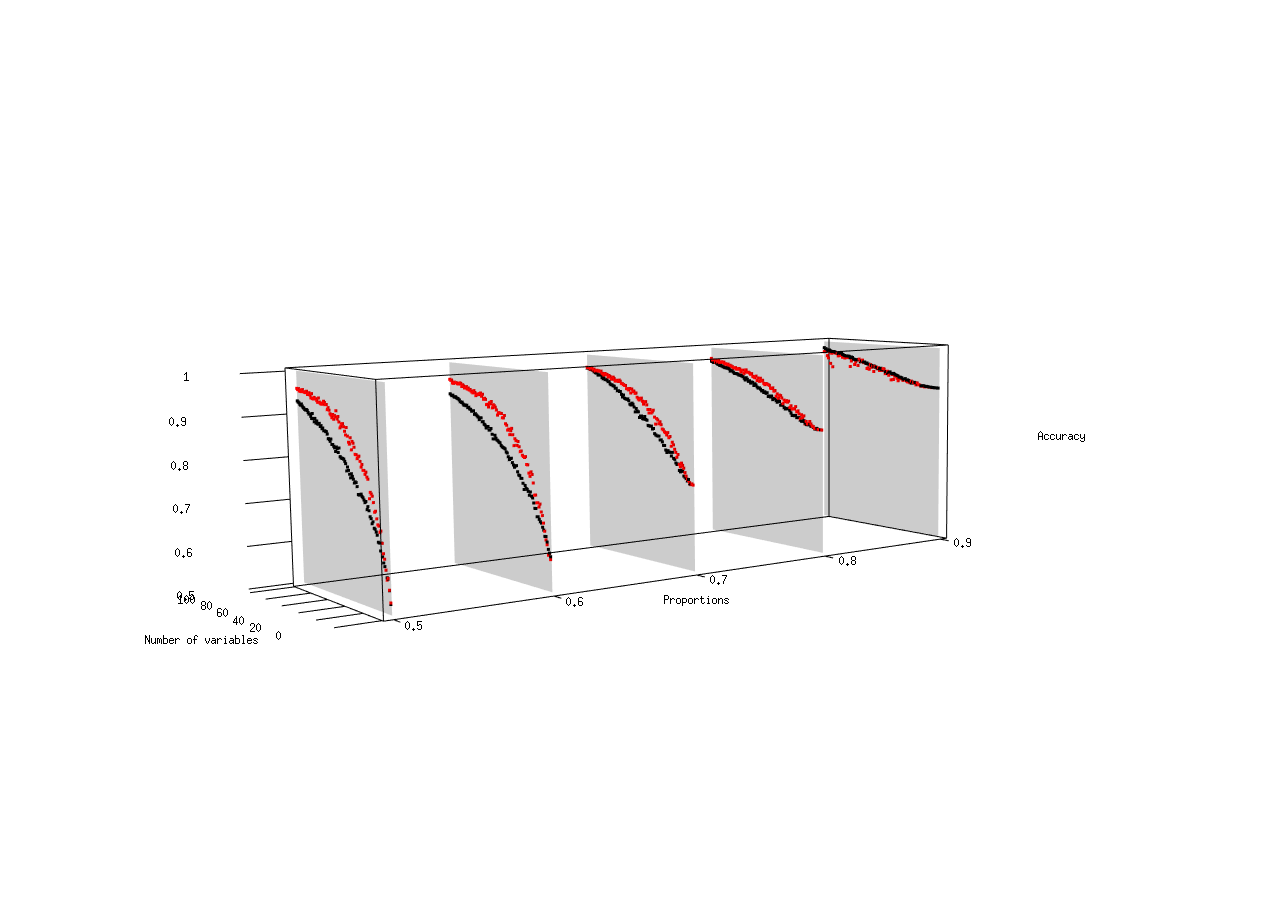} \\ (a) Accuracy
\\\includegraphics[trim=2cm 10cm 2cm 11.5cm,clip=true,width=\textwidth]{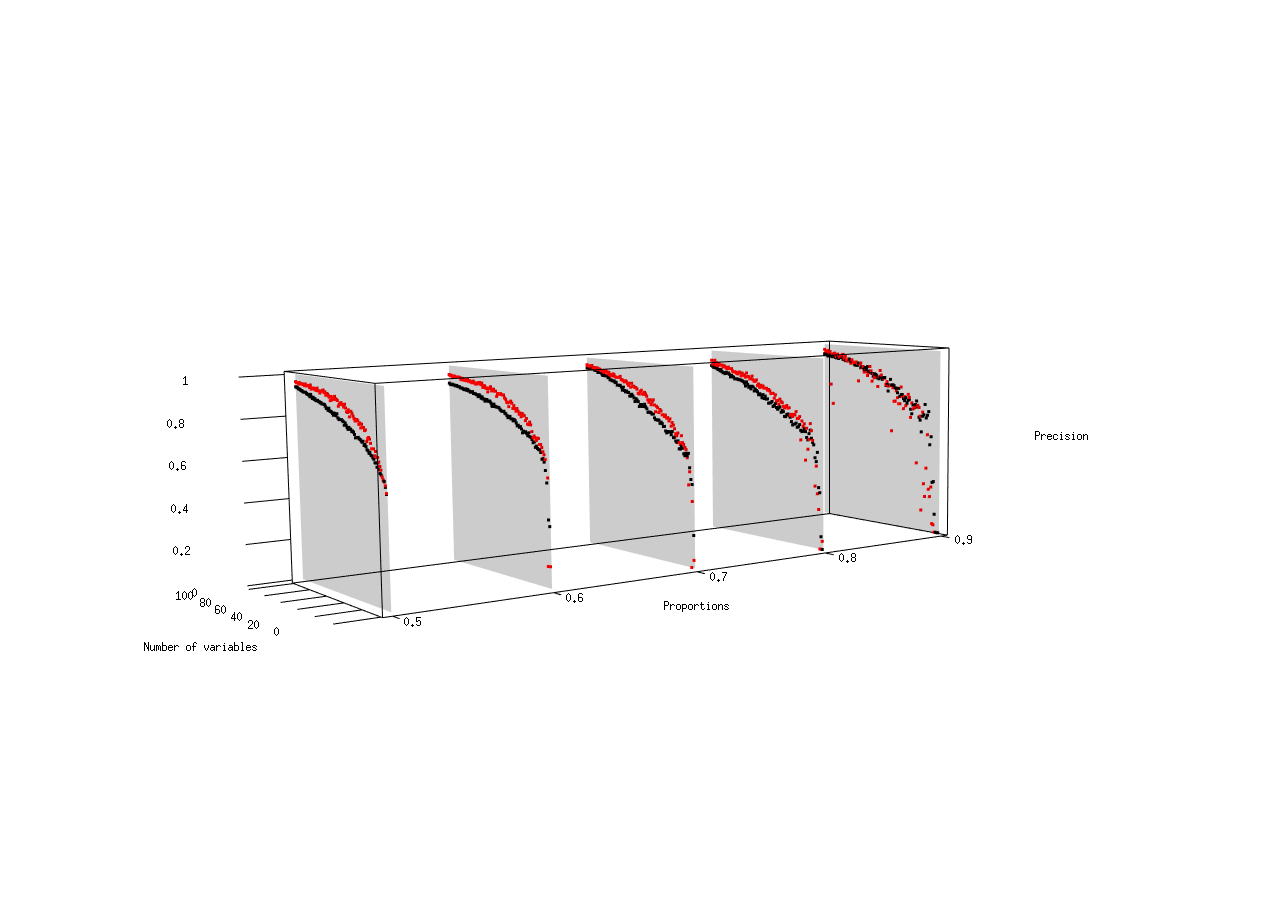}\\ (b) Precision\\
\includegraphics[trim=2cm 10cm 2cm 11.5cm,clip=true,width=\textwidth]{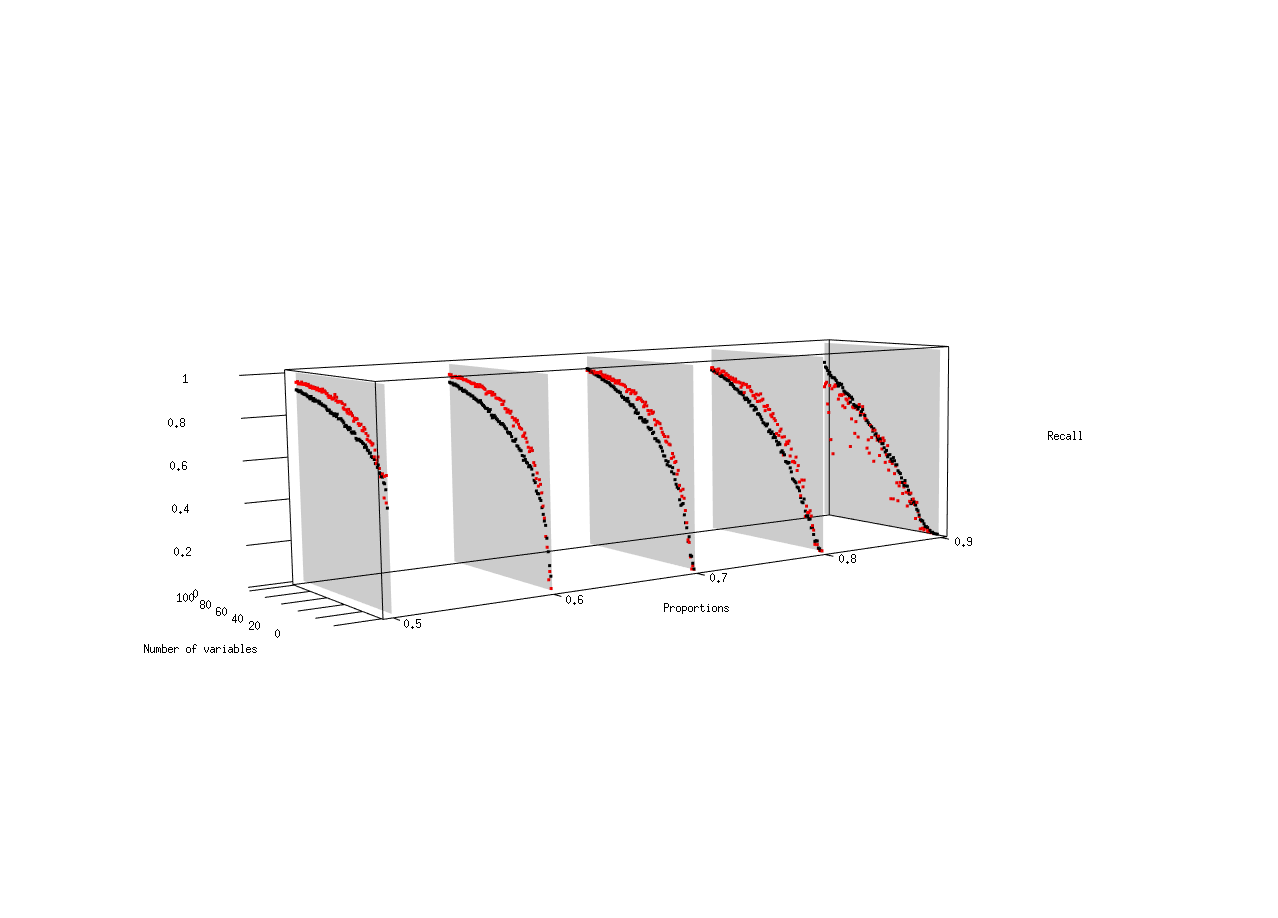}\\ (c) Recall
\\\includegraphics[trim=2cm 10cm 2cm 11.5cm,clip=true,width=\textwidth]{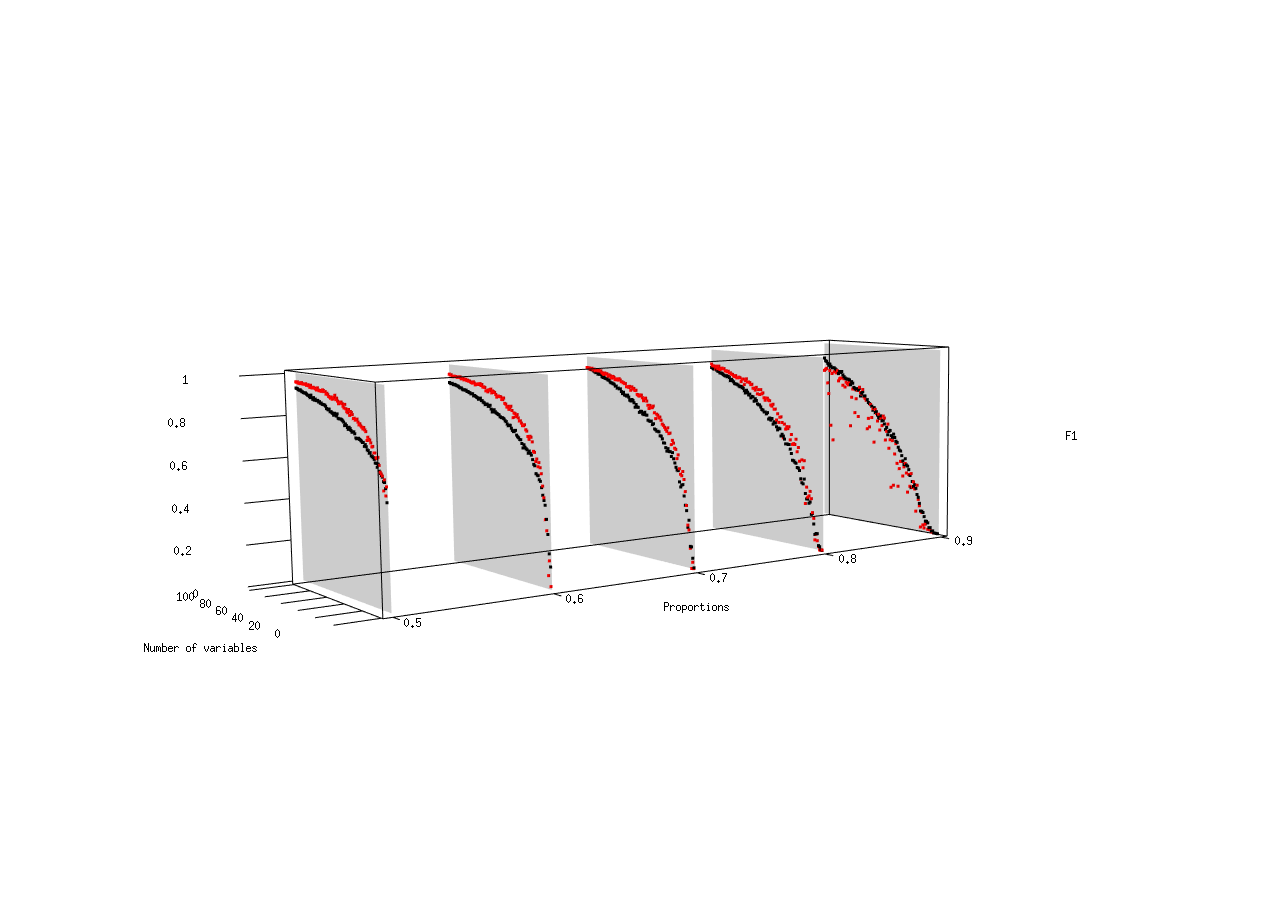} \\ (d) F1\\
\caption{The mean accuracy (a), precision (b), recall (c) and F1 (d) scores per number of included variables and $p_T$ over 20 simulations when features are discrete. It can be seen that only when $p_T = 0.9$, the discrete choice model (black dots) performs better than the artificial neural network (red dots). For the other proportions, the neural network generates better predictions. It can also be seen that more variables results usually in better predictions exception for the neural network when $p_T = 0.9$.}
\end{center}
\label{fig:ScoresDiscreteImbalanced}
\end{figure}

\begin{figure}
\begin{center}
\includegraphics[trim=2cm 9cm 2cm 10cm,clip=true,width=0.99\textwidth]{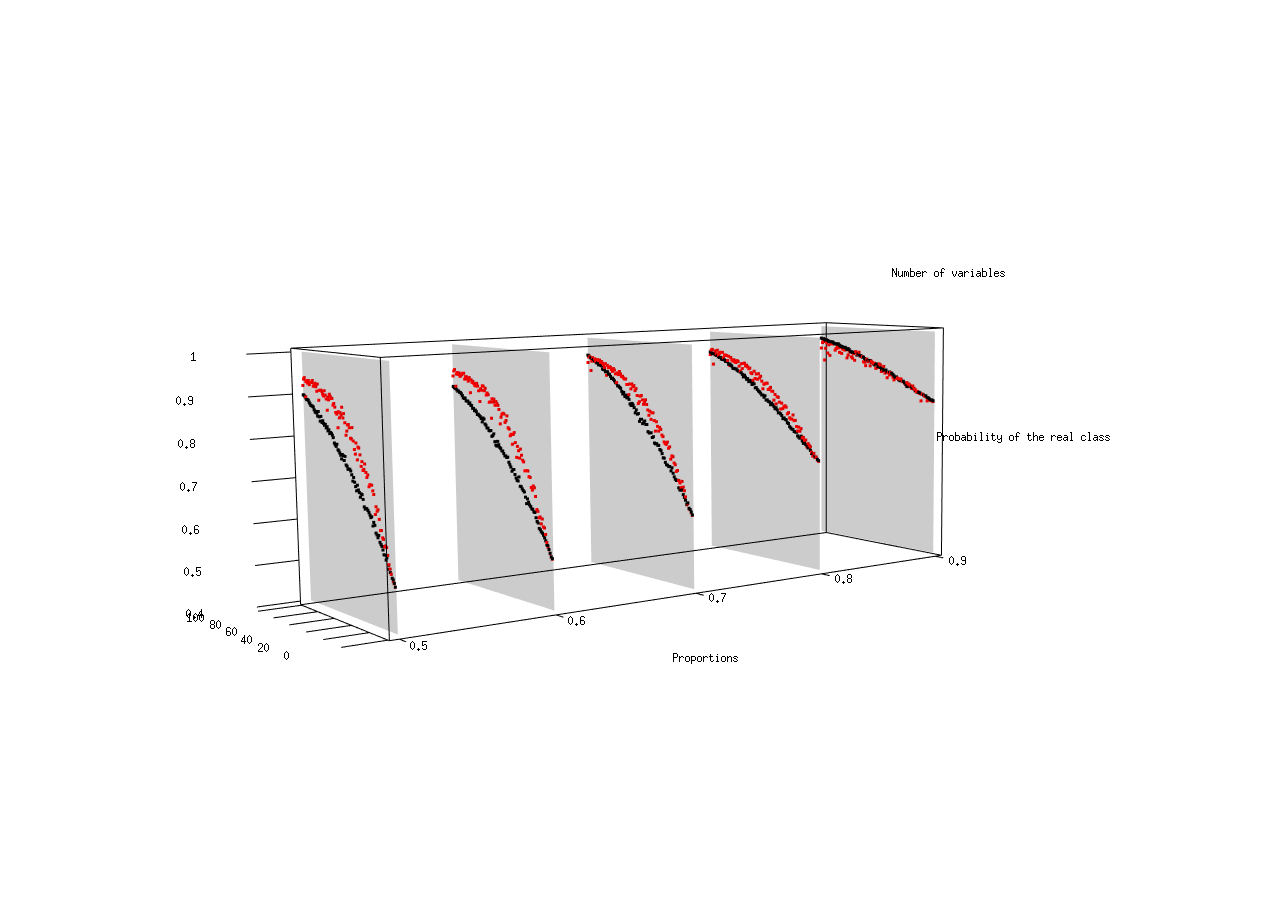}\\ (a) actual  \vspace{2cm}\\ \includegraphics[trim=2cm 9cm 2cm 10cm,clip=true,width=0.99\textwidth]{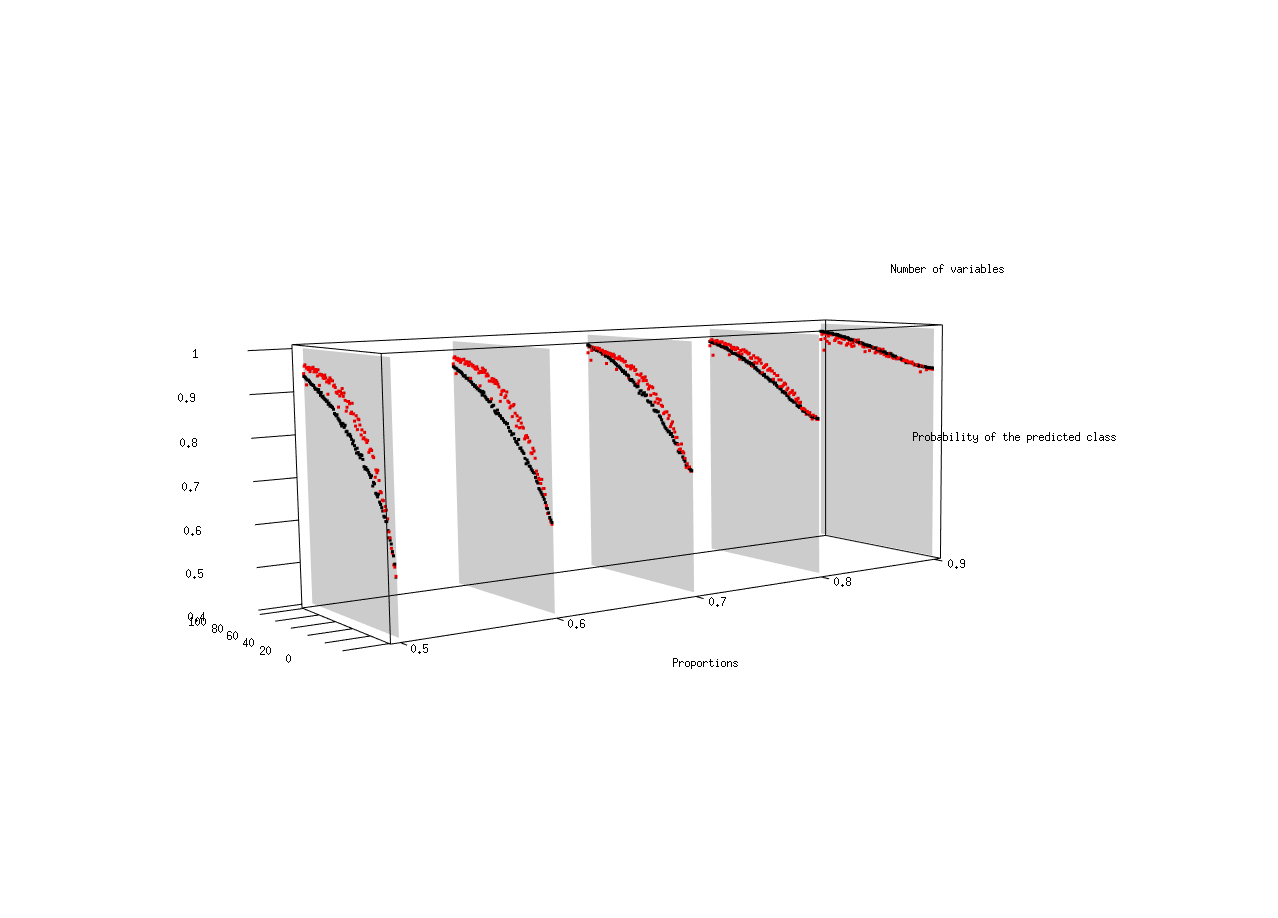} \\(b) predicted
\caption{Probabilities of the actual (a) and predicted (b) classes generated by the artificial neural network (red) and the discrete choice model (black) for continuous variables. The artificial neural network has more confidence into its predictions, since the probabilities are higher (panel b) except for $p_T = 0.9$. In addition, the same can be seen for the predicted probabilities of the actual class. When $p_T=0.9$, there is no clear winner, even though the DC seems to be more consistent.}
\end{center}
\label{fig:ProbaDiscreteImbalanced}

\end{figure}

\clearpage
\bibliographystyle{te}
\bibliography{refs_dc_ann}

\end{document}